\documentclass[10pt, a4paper]{article}
\usepackage{lrec2026} % this is the new style

\usepackage{latexsym}
\usepackage{pgfplots}
\pgfplotsset{compat=1.18}
\usepackage{amsmath}
\usepackage{pgfplotstable}
\usepackage{tikz}
\usepackage{afterpage}
\usepackage{tabularx}
\usepackage{booktabs}
\usepackage{float}
\usepackage{graphicx}
\usepackage{tcolorbox}
\usepackage{xcolor}
\usepackage{multicol}

\tcbuselibrary{skins, breakable}
\usepackage{multirow}
\usepackage[table]{xcolor}
\usepackage{hyperref}
\usepackage{url}

\usepackage{fontspec}
\usepackage{polyglossia}
\usepackage{enumitem}

\setmainlanguage{english}
\setotherlanguage{persian}

\newfontfamily\persianfont[Script=Arabic,Scale=1.0]{BNazanin.ttf}

\newenvironment{persianpar}
{\begin{otherlanguage}{persian}\persianfont\rightskip=0pt plus 1fil\leftskip=0pt plus 1fil\parindent=0pt}
{\par\end{otherlanguage}}

\newtcolorbox{persianbox}[1][]{
  colback=gray!5,
  colframe=gray!40,
  boxrule=0.5pt,
  rounded corners,
  fontupper=\persianfont,
  #1
}

\definecolor{mybg}{HTML}{F1F9FF}
\definecolor{myaccent}{HTML}{095D7E}
\definecolor{myborder}{HTML}{14967F}
\definecolor{mylight}{HTML}{E2FCD6}
\definecolor{mysoft}{HTML}{CCECEE}
\definecolor{mybg2}{HTML}{E8F4FF}
\definecolor{myaccent2}{HTML}{1B6B85}
\definecolor{myborder2}{HTML}{0F8B7A}
\definecolor{mylight2}{HTML}{D8F0E8}
\definecolor{metameta}{HTML}{FFF5E6}
\definecolor{metaborder}{HTML}{D4843C}
\definecolor{errorcolor}{HTML}{8B4513}
\definecolor{successcolor}{HTML}{2F7D32}
\definecolor{warningcolor}{HTML}{F57C00}
\definecolor{infocolor}{HTML}{1976D2}

\title{PersianMedQA: Evaluating Large Language Models on a Persian-English Bilingual Medical Question Answering Benchmark}

\name{
	\small\bfseries
	Mohammad Javad Ranjbar Kalahroodi$^{1}$, 
	Sepehr Karimi$^{1}$\textsuperscript{*}, 
	Amirhossein Sheikholselami$^{1}$\textsuperscript{*}, \\
	\small\bfseries
	Sepideh Ranjbar Kalahroodi$^{2}$, 
	Heshaam Faili$^{1}$, 
	Azadeh Shakery$^{1,3}$
}

\address{%
$^{1}$School of Electrical and Computer Engineering, University of Tehran, Iran\\
$^{2}$Shahid Beheshti University of Medical Sciences, Iran\\
$^{3}$Institute for Research in Fundamental Sciences (IPM), Tehran, Iran\\
\texttt{\{mohammadjranjbar, ah.sheikh, sepehrkarimi, hfaili, shakery\}@ut.ac.ir}}

\address{
$^{1}$School of Electrical and Computer Engineering, University of Tehran, Iran\\
$^{2}$Shahid Beheshti University of Medical Sciences, Iran\\
$^{3}$Institute for Research in Fundamental Sciences (IPM), Tehran, Iran\\
\texttt{\{mohammadjranjbar, ah.sheikh, sepehrkarimi, hfaili, shakery\}@ut.ac.ir}\\\textsuperscript{*}These authors contributed equally to this work and are listed in random order.}

\abstract{
Large Language Models (LLMs) have achieved remarkable performance on a wide range of Natural Language Processing (NLP) benchmarks, often surpassing human-level accuracy. However, their reliability in high-stakes domains such as medicine, particularly in low-resource languages, remains underexplored. In this work, we introduce PersianMedQA, a large-scale dataset of 20,785 expert-validated multiple-choice Persian medical questions from 14 years of Iranian national medical exams, spanning 23 medical specialties and designed to evaluate LLMs in both Persian and English. We benchmark 41 state-of-the-art models, including general-purpose, Persian, and medical LLMs, in zero-shot and chain-of-thought (CoT) settings. Our results show that closed-weight general models (e.g., GPT-4.1) consistently outperform all other categories, achieving 83.09\% accuracy in Persian and 80.7\% in English, while Persian LLMs such as Dorna underperform significantly (e.g., 34.9\% in Persian), often struggling with both instruction-following and domain reasoning. We also analyze the impact of translation, showing that while English performance is generally higher, 3-10\% of questions can only be answered correctly in Persian due to cultural and clinical contextual cues that are lost in translation. Finally, we demonstrate that model size alone is insufficient for robust performance without strong domain or language adaptation. PersianMedQA provides a foundation for evaluating bilingual and culturally grounded medical reasoning in LLMs. The dataset, along with a bilingual medical dictionary, is publicly available at \href{https://huggingface.co/datasets/MohammadJRanjbar/PersianMedQA}{PersianMedQA}.
\\ \newline }

\begin{document}

\maketitleabstract

\section{Introduction}
\label{sec:introduction}

Large Language Models (LLMs) have become the go-to solution for many tasks, showcasing promising results on standard benchmarks, potentially replacing humans across various domains~\cite{brown2020languagemodelsfewshotlearners, openai2024gpt4technicalreport}. However, their reliability in tasks that require real attention to detail, such as those directly impacting human life, remains concerning~\cite{bommasani2022foundation}. Medical tasks, such as clinical decision-making, represent a critical domain where experts must possess comprehensive knowledge in cultural contexts, medical principles, pharmaceutical information, and numerous other specialized areas within healthcare. In other words, clinical excellence requires more than just biomedical knowledge~\cite{campinha2002process}.

\begin{figure}[t]
	\centering
	\begin{tcolorbox}[
		enhanced,
		colback=mybg,
		colframe=myborder2,
		arc=3mm,
		boxrule=1pt,
		title={\scriptsize\bfseries Medical Examples},
		fonttitle=\scriptsize\bfseries\centering,
		left=3pt, right=3pt, top=2pt, bottom=2pt
		]
		\scriptsize
		\textbf{\textcolor{myaccent}{Clinical:}}
		\begin{tcolorbox}[
			colback=mylight2,
			colframe=myaccent2,
			boxrule=0.5pt,
			arc=2mm,
			left=2pt, right=2pt, top=2pt, bottom=2pt
			]
			\scriptsize
			A 48-year-old man presents with chest pain (4h), anterior ST-elevation, sweating, BP 90/60 mmHg, distended neck veins, and basal rales. Most effective treatment?\\[2pt]
			1. Fibrinolytic + emergency angioplasty if needed\\
			2. Fibrinolytic only\\
			3. Emergency angioplasty\\
			4. Fibrinolytic + angioplasty after 48h\\[2pt]
			\textbf{Answer:} \textcolor{successcolor}{3}
		\end{tcolorbox}
		\vspace{0.15cm}
		\textbf{\textcolor{myaccent}{Non-Clinical:}}
		\begin{tcolorbox}[
			colback=mylight2,
			colframe=myborder2,
			boxrule=0.5pt,
			arc=2mm,
			left=2pt, right=2pt, top=2pt, bottom=2pt
			]
			\scriptsize
			All of the following can cause acute retinal necrosis, \textit{except}:\\[2pt]
			1. Cytomegalovirus\\
			2. Herpes simplex type 1\\
			3. Toxoplasmosis\\
			4. Varicella Zoster\\[2pt]
			\textbf{Answer:} \textcolor{successcolor}{3}
		\end{tcolorbox}
	\end{tcolorbox}
	\caption{A translated medical question example from the dataset.}
	\label{fig:dataset-examples}
\end{figure}

\begin{figure*}[t!]
	\centering
	\includegraphics[width=0.75\textwidth]{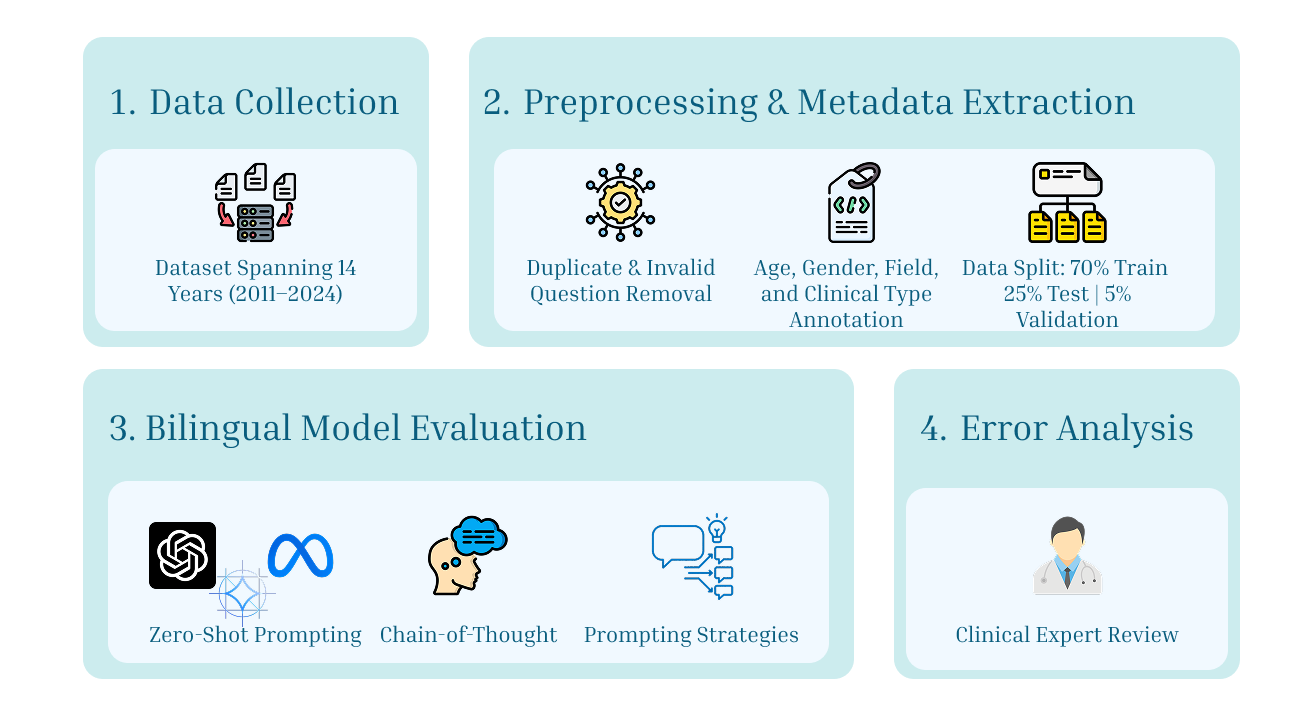}
	\caption{Overview of the PersianMedQA dataset construction process, including data collection, cleaning, annotation, and partitioning steps.}
	\label{fig:paper-overview}
\end{figure*}

Although recent works have demonstrated that LLMs may achieve high accuracy on English medical question-answering tasks~\cite{singhal2022largelanguagemodelsencode,saab2024capabilitiesgeminimodelsmedicine}, their performance drops significantly in other languages \cite{QIN2025101118, Alonso_2024}. This gap is particularly pronounced in medicine, where high-quality corpora are centered on English, restricting the models' applicability in global healthcare settings. Importantly, simply translating questions is inadequate, as such pipelines can strip away critical terminology, subtle cultural cues, and localized standards of care, potentially leading to life-threatening consequences in clinical practice \cite{10.1145/3531146.3533244}.

Medical practice is inherently shaped by contextual factors, including sociocultural, socioeconomic, regional, and healthcare system variables that extend beyond language translation \cite{KLEINMAN197885, betancourt2003defining}. Clinical decision-making protocols and symptom interpretation vary significantly across healthcare systems and populations due to genetic variations, dietary patterns, climate-related health risks, and socioeconomic determinants, with the same clinical presentation potentially indicating different underlying pathologies across ethnic groups \cite{kirmayer2001cultural,wennberg2002unwarranted, risch2002categorization, zborowski1952cultural}. Additionally, vaccination schedules, drug availability, and standard-of-care protocols differ markedly between regions, making direct translation of medical guidelines inefficient. These considerations highlight why medical AI systems cannot rely solely on linguistic translation but must incorporate an understanding of regional medical practices and population-specific health patterns.

These contextual complexities are particularly pronounced in low-resource language settings, where the intersection of linguistic barriers and distinct medical practices creates compounded challenges for AI evaluation. Limited research has investigated the specific factors that mislead LLMs in medical contexts, particularly in multilingual and low-resource language settings like Persian. A deeper investigation into the medical sub-fields in which LLMs excel or underperform is essential for identifying suitable use cases and implementing necessary safeguards.

To fill this gap, we introduce PersianMedQA, a large-scale, expert-annotated dataset covering 23 medical specialties. Given the scarcity of standardized Persian medical terminology resources, the dataset includes a comprehensive bilingual dictionary of Persian medical terms for consistent terminology usage during evaluation and model adaptation. As a benchmark, we evaluate state-of-the-art models, including general-purpose models, Persian LLMs, and medical LLMs on both original Persian questions and their English translations. Throughout this work, we use \emph{medical LLMs} to refer to models that have undergone domain adaptation for medicine, either via domain-adaptive pre-training on biomedical corpora or supervised instruction fine-tuning on medical tasks. Similarly, \emph{Persian LLMs} refers to models domain-adapted for Persian, via continued pre-training or Persian instruction fine-tuning. Our experiments uncover a substantial language gap: closed-weight models such as GPT-4.1 significantly outperform open-weight counterparts. Notably, Persian LLMs exhibited minimal understanding of the Persian medical field and performed the worst, while medical LLMs showed only modest improvements and failed to generalize effectively to Persian clinical data. Figure~\ref{fig:paper-overview} illustrates the overall workflow of our study.

Section~\ref{sec:related} reviews prior work on medical QA benchmarks and Persian language models. Section~\ref{sec:construction} describes the PersianMedQA dataset construction. Section~\ref{sec:experiments} presents our experimental setup and evaluations. Section~\ref{sec:conclusion} concludes with key findings and future research directions.

\section{Related Work}
\label{sec:related}

\subsection{Medical Question Answering Datasets and Multilingual Challenges}

Medical question answering has emerged as a critical benchmark for evaluating machine reasoning capabilities in high-stakes healthcare domains. The field has evolved from early information retrieval benchmarks~\cite{ATHENIKOS20101, cao2011askhermes} to standardized datasets such as \textsc{PubMedQA}~\cite{jin2019pubmedqadatasetbiomedicalresearch}, \textsc{MedQA}~\cite{jin2020diseasedoespatienthave}, and \textsc{MedMCQA}~\cite{pal2022medmcqalargescalemultisubject}, driving domain-specific model development like \textsc{BioBERT}~\cite{Lee_2019} and \textsc{PubMedBERT}~\cite{Gu_2021}. However, most benchmarks focus exclusively on English, creating significant evaluation gaps. While native-language datasets have emerged—including CMB~\cite{wang2024cmbcomprehensivemedicalbenchmark}, Huatuo-26M~\cite{li2023huatuo26mlargescalechinesemedical}, MedQA-SWE~\cite{hertzberg-lokrantz-2024-medqa}, FrenchMedMCQA~\cite{labrak-etal-2022-frenchmedmcqa}, and HeadQA~\cite{vilares-gomez-rodriguez-2019-head}—many frameworks rely on problematic ``translate-test'' methodologies~\cite{jin2023betteraskenglishcrosslingual, liu2025translationneedstudysolving} (the practice of machine-translating non-English inputs into English before evaluation, rather than evaluating in the source language) that distort clinical terminology and miss culturally-specific practices~\cite{jin2023betteraskenglishcrosslingual,liu2025translationneedstudysolving}. Recent multilingual efforts like \textsc{MedExpQA}~\cite{Alonso_2024} show around 10\% accuracy drops for non-English languages, with critical gaps remaining for low-resource languages requiring native evaluation approaches.

\subsection{LLMs in Medical Practice}

LLMs have transformed medical AI applications, with specialized models demonstrating remarkable capabilities on standardized medical examinations. \textsc{Med-PaLM 2}~\cite{singhal2023expertlevelmedicalquestionanswering} achieved groundbreaking performance on the USMLE, while general-purpose models like \textsc{GPT-4}~\cite{openai2024gpt4technicalreport} showed impressive zero-shot performance across medical QA benchmarks~\cite{nori2023capabilitiesgpt4medicalchallenge}. Recent advances include open-weight models such as \textsc{MEDITRON-70B}~\cite{chen2023meditron70bscalingmedicalpretraining}, multilingual approaches like \textsc{MMed-Llama 3}~\cite{qiu2024buildingmultilinguallanguagemodel} covering six languages, and specialized Chinese models such as \textsc{TCMChat}~\cite{DAI2024107530} and \textsc{BianCang}~\cite{wei2024biancangtraditionalchinesemedicine}. However, systematic evaluation across diverse languages and clinical settings remains limited, particularly for morphologically rich and low-resource languages such as Persian.

\subsection{Persian Language Models and Medical Applications}

Persian natural language processing has witnessed significant progress with the development of robust monolingual models. \textsc{ParsBERT}~\cite{Farahani_2021} established strong baselines for various Persian NLP tasks, consistently outperforming multilingual alternatives on sentiment analysis and text classification benchmarks. Recent advances include \textsc{Dorna}~\cite{partai_dorna2_llama3.1_8b_instruct}, a large-scale Persian language model. In the medical domain, \textsc{SINA-BERT}~\cite{taghizadeh2021sinabertpretrainedlanguagemodel} represents an early attempt at Persian medical NLP, utilizing pre-training on large-scale medical corpora including both formal and informal medical texts from diverse online resources. Furthermore, existing Persian medical NLP efforts lack the expert validation and standardized evaluation protocols necessary for reliable clinical assessment, highlighting the need for comprehensive Persian medical QA benchmarks with rigorous validation procedures.
\begin{figure}[t]
	\centering
	\includegraphics[width=0.48\textwidth]{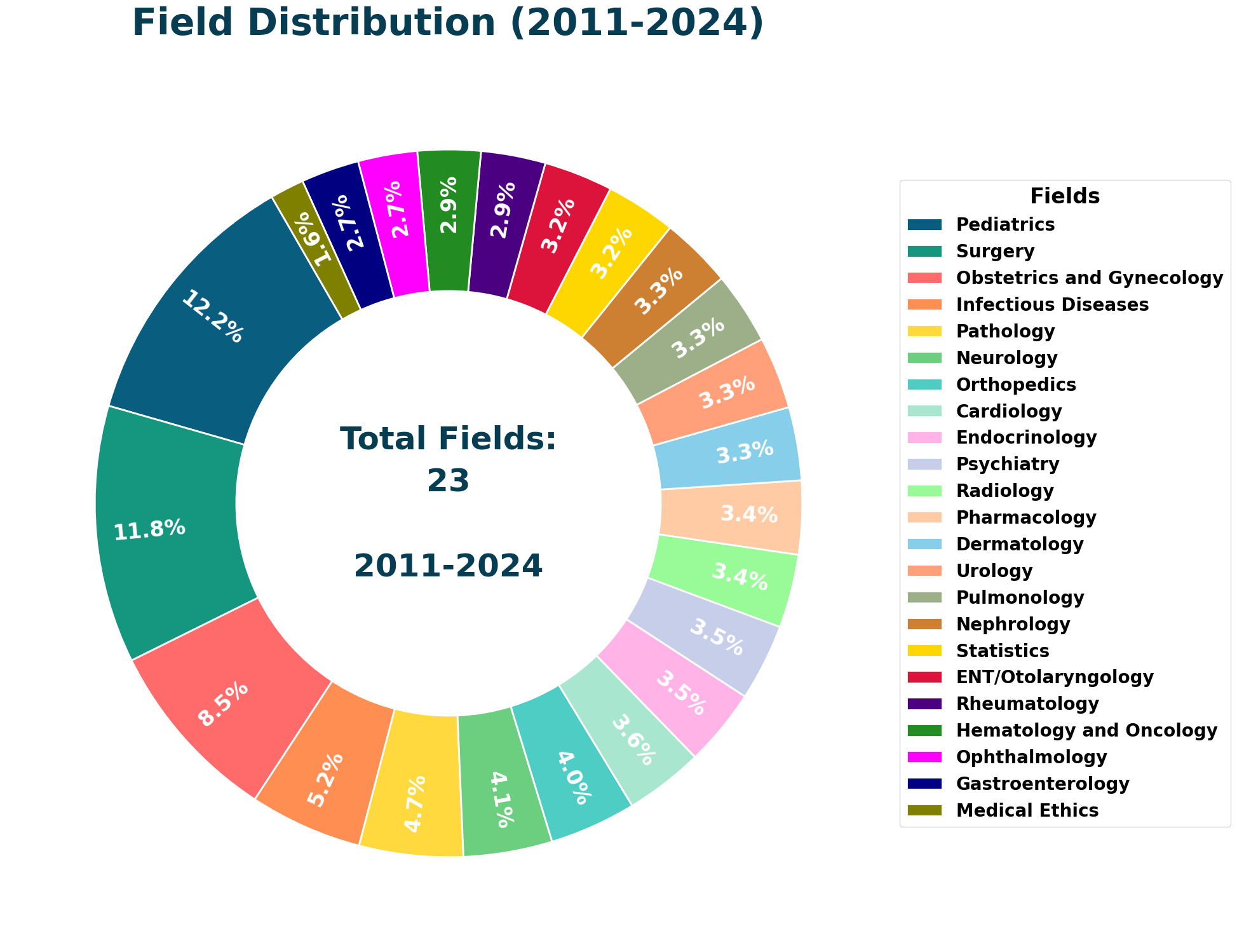}
	\caption{Distribution of medical fields in the dataset.}
	\label{fig:dataset-pie}
\end{figure}

\section{PersianMedQA Construction}
\label{sec:construction}

The PersianMedQA dataset was developed by collecting 14 years of multiple-choice questions from the official Iranian medical residency and pre-residency examinations administered by the Medical Education Assessment Center (Sanjeshp) under the Iranian Ministry of Health, the governmental body responsible for national standardized examinations in the medical domain, analogous to organizations such as the National Board of Medical Examiners (NBME) in the United States. These exams serve as the mandatory licensing gateway for medical graduates seeking specialist training in Iran, ensuring that all questions reflect real-world, high-stakes clinical evaluation standards at the national level. Each exam was created by the official Iranian medical board and reflects real-world, high-stakes evaluation standards. Each item includes the question text, four answer options, the correct answer key, and the medical field to which the question belongs. Figure~\ref{fig:dataset-examples} presents representative examples of clinical and non-clinical questions. The raw dataset underwent a rigorous preprocessing pipeline to ensure quality, consistency, and relevance for multilingual medical QA evaluation.

\subsection{Data Cleaning and Filtering}

In order to eliminate noise and redundancy, we ran a three-step cleaning pipeline:
\begin{itemize}
    \item \textbf{Duplicate Removal:} Automatically prune exact and near-duplicate questions using string matching and sentence-embedding similarity from the Language-agnostic BERT Sentence Embedding (LaBSE) model \cite{feng2020labse} to maintain diversity.
    \vspace{-5 pt}

    \item \textbf{Image Dependent Exclusion:} Discard any question that relies on medical images (e.g., radiographs, histology slides) so the benchmark remains purely text-based.
    \vspace{-5 pt}

    \item \textbf{Answer Key Verification:} All reference answers were provided by the official Iranian Medical Board and underwent a rigorous three-level verification process by the National Center for Medical Education Assessment (Sanjesh): (1) Initial review by board-certified medical professionals, (2) Public comment period where students and practitioners could report concerns or discrepancies, and (3) Final review and correction incorporating community feedback. In addition to this official validation, we manually removed questions with incorrect answer formatting (e.g., missing or mismatched option numbers) and those with multiple plausible correct answers to ensure evaluation reliability.
\end{itemize}

\begin{figure*}[t]
	\centering
	\includegraphics[width=0.85\textwidth]{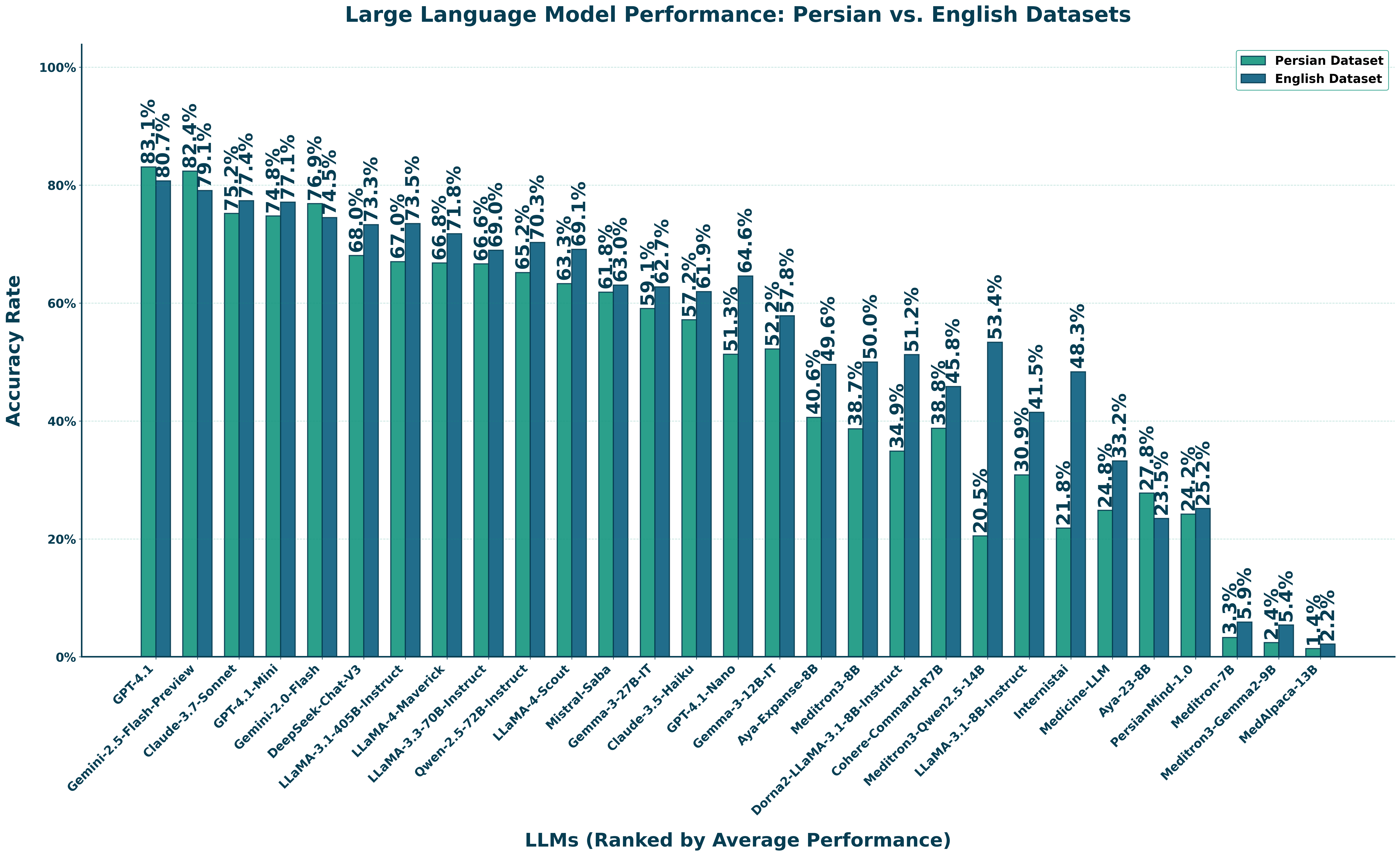}
	\caption{Overall accuracy of models on Persian and English test sets.}
	\label{fig:overall-accuracy}
\end{figure*}

\subsection{Annotation and Categorization}

To enhance interpretability and analysis, the cleaned dataset was annotated as follows:
\begin{itemize}
\item \textbf{Subject Verification:} Most questions already contained subject tags (i.e., the medical specialty labels shown in Figure~\ref{fig:dataset-pie}, such as cardiology, surgery, or pediatrics) from the original examination. For questions lacking subject tags, both medical specialists and Gemini 2.5-Flash independently classified them, achieving over 90\% agreement. Final subject labels were determined through expert medical review to ensure high accuracy.
\vspace{-5 pt}

\item \textbf{Domain Classification:} Questions were labeled as \emph{clinical} (patient cases and diagnosis) or \emph{non-clinical} (basic sciences and theoretical concepts). This classification was performed using Gemini 2.5-Flash and validated by a medical specialist.
\vspace{-5 pt}

\item \textbf{Demographic Extraction:} We utilized Gemini 2.5-Flash to automatically extract patient attributes (e.g., age, gender) for every question, motivated by the need to analyze  data distribution to ensure comprehensive representation across patient demographics and enable future research on potential LLM performance gaps in specific demographic subgroups.
\end{itemize}
\vspace{-5 pt}

All medical annotations and validations were performed by a board-certified internal medicine specialist with 5 years of clinical practice (see Appendix~\ref{appendix:specialist} for full details).

\subsection{Dataset Overview}

The \textbf{PersianMedQA} dataset comprises 20,785 unique, expert-validated multiple-choice medical questions, collected over 14 years from Iranian national residency and pre-residency exams. Approximately 70\% of the questions are classified as clinical, with the remaining 30\% labeled as non-clinical. The items span 23 medical specialties, covering a broad range of topics relevant to medical education and practice.

We partitioned PersianMedQA into training (14,549), validation (1,000), and test (5,236) splits using stratified sampling based on year and field to facilitate future research, including potential fine-tuning studies, and to establish standardized evaluation protocols. Figure~\ref{fig:dataset-pie} summarizes the distribution of questions across medical domains.

\textbf{Demographic Coverage:} To ensure comprehensive representation and enable future bias analysis, we systematically extracted patient demographics using LLM-based extraction validated by medical experts. Table~\ref{tab:demographics_summary} presents the distribution across gender, age categories, and question types. The dataset demonstrates balanced representation across demographics, with substantial coverage of both male and female patients, diverse age groups from infants to adults, and a clinically-relevant mix of case-based and theoretical questions.

\begin{table}[h]
\centering
\caption{Demographic distribution in PersianMedQA dataset.}
\label{tab:demographics_summary}
\scriptsize
\begin{tabular}{llr}
\toprule
\textbf{Category} & \textbf{Subcategory} & \textbf{Count} \\
\midrule
\multirow{3}{*}{Gender} 
& Male & 5,590 \\
& Female & 5,831 \\
& Unspecified & 9,361 \\
\midrule
\multirow{4}{*}{Age} 
& Infant (0--1) & 1,101 \\
& Child (2--17) & 2,675 \\
& Adult (18+) & 10,241 \\
& Unspecified & 6,765 \\
\midrule
\multirow{2}{*}{Question Type} 
& Clinical & 14,724 \\
& Non-Clinical & 6,061 \\
\bottomrule
\end{tabular}
\end{table}

\subsection{Data Contamination and Evaluation Integrity}

To ensure the reliability of our medical evaluation, we implemented multiple safeguards against data contamination and memorization artifacts:

\textbf{Secure Sourcing}: The dataset is not from easily crawlable, free public websites. Questions come from official Iranian medical residency exams administered by Sanjeshp in PDF format, providing an additional layer of protection against training data leakage.

\textbf{Exact Search}: We conducted exact match searches on a randomly sampled subset of under 50 questions, querying both full question stems and isolated key medical terms across publicly accessible web sources. The analysis revealed \textbf{minimal overlap}, with the vast majority of sampled questions returning no verbatim matches on any publicly accessible platform, indicating \textbf{limited or no data leakage} into LLM training corpora.

\begin{figure}[h]
\centering
\includegraphics[width=0.48\textwidth]{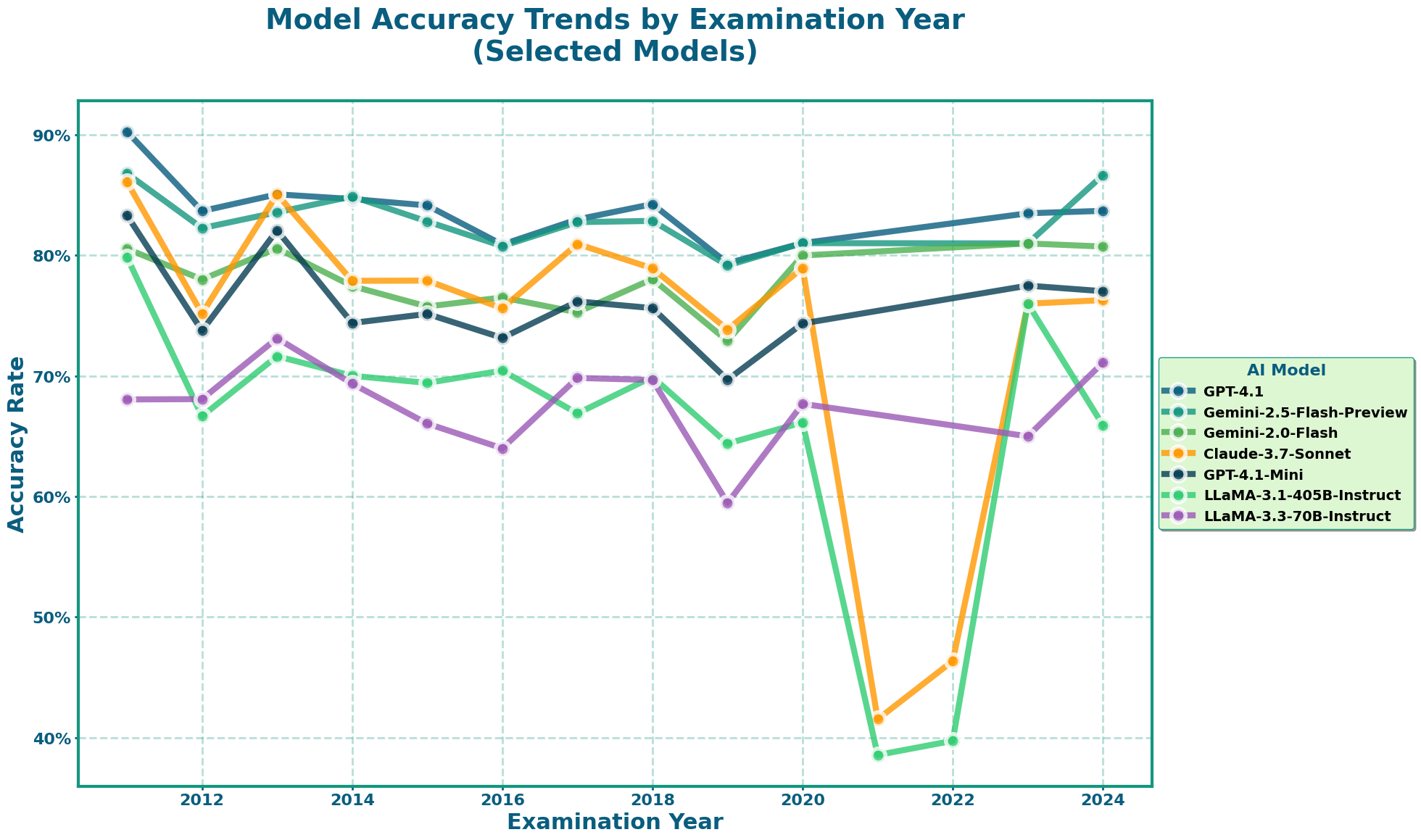}
\caption{LLM performance across exam years (2011-2024).}
\label{fig:temporal-analysis}
\end{figure}

\textbf{Temporal Analysis}: To empirically verify the absence of data leakage, we performed a year-by-year accuracy analysis across examinations from 2011-2024 (Figure~\ref{fig:temporal-analysis}). Model performance remained consistent even on 2023-2024 questions that post-date training cutoffs for most evaluated models, strongly suggesting our dataset was not present in training corpora. The observed performance decrease in 2020-2021 likely reflects increased exam difficulty during the COVID-19 pandemic period rather than data inconsistencies, as students had extended study time and examiners reportedly created more challenging assessments.

\subsection{English Translation}
To enable bilingual evaluation, we generated English translations of the dataset using three methods: Google Translate, the GPT-4.1 API, and the Gemini-2.5-Flash API. All translations were assessed by our board-certified medical expert for accuracy of medical terminology and preservation of clinical meaning. Both GPT-4.1 and Gemini-2.5-Flash produced more accurate, natural translations than Google Translate, with expert validation confirming superior preservation of medical concepts and terminology. Due to its combination of quality, accessibility, and expert-validated accuracy, we use Gemini-2.5-Flash translations as our default in all subsequent experiments.

\section{Experiments}
\label{sec:experiments}

\subsection{Zero-shot Scenario}

We conducted zero-shot evaluations on the PersianMedQA dataset using a wide range of state-of-the-art open-weight and closed-weight LLMs in both Persian and English. All models were evaluated with temperature set to 0 and allowed to generate up to their maximum generation length. We employed identical prompts and evaluation protocols across all models to ensure fair comparison. Figure~\ref{fig:zeroshot_prompt} shows the complete prompt template used for all zero-shot evaluations. We used English prompts across both Persian and English evaluation settings, as most models demonstrated superior instruction-following capabilities in English compared to Persian. All results reported in this study are based on the 5,236-question test set to facilitate reproducible evaluation and reduce computational costs. We validated this approach by running experiments on multiple models using the full dataset, confirming that results were consistent with the smaller test set. 

\begin{figure}[htbp]
\centering
\begin{tcolorbox}[
breakable,
enhanced,
colback=mybg,
colframe=myborder2,
arc=3mm,
boxrule=1pt,
title={\scriptsize\bfseries Zero-shot Prompt},
fonttitle=\scriptsize\bfseries\centering,
left=4pt, right=4pt, top=4pt, bottom=4pt
]
\scriptsize
You are a medical expert tasked with answering multiple-choice medical questions.

\vspace{0.3cm}
\textbf{\textcolor{myaccent}{Question Format:}}
\begin{tcolorbox}[
colback=mylight2,
colframe=myaccent2,
boxrule=0.5pt,
arc=2mm,
left=3pt, right=3pt, top=3pt, bottom=3pt
]
\scriptsize
\begin{verbatim}
Question: [Medical question text]
1: [Option 1]
2: [Option 2]
3: [Option 3]
4: [Option 4]
\end{verbatim}
\end{tcolorbox}
\vspace{0.3cm}
\textbf{\textcolor{myaccent}{Important Notes:}}
\begin{itemize}[nosep, leftmargin=10pt]
  \item Select the best answer from the provided choices.
  \item Your output must be \textbf{only the option number} (1, 2, 3, or 4).
  \item Do \textbf{not} add explanations or extra text.
  \item Base your answers on authoritative medical knowledge.
\end{itemize}
\end{tcolorbox}
\caption{The zero-shot prompt used for evaluation}
\label{fig:zeroshot_prompt}
\end{figure}

\begin{figure*}[t]
    \centering
    \includegraphics[width=0.85\textwidth]{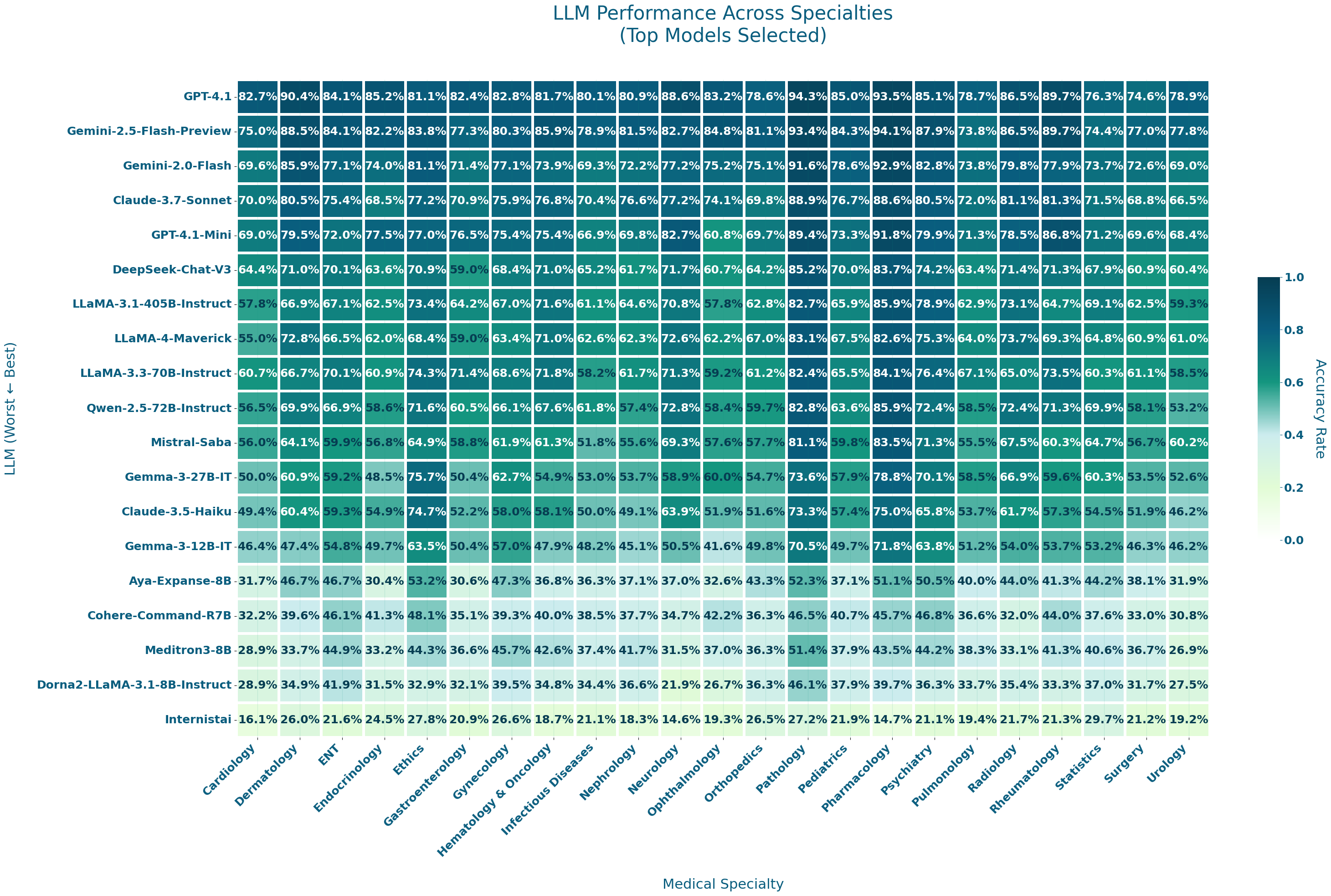} 
    \caption{Heatmap showing the accuracy of each model across all medical specialties in the PersianMedQA dataset. Each cell represents the accuracy for a particular model-field pair. The full model list is available in Appendix~\ref{appendix:overall-performance}.}
    \label{fig:field-heatmap}
\end{figure*}

Figure~\ref{fig:overall-accuracy} presents the overall accuracy of the evaluated models on both Persian and English test sets. Among all models, the closed-weight GPT-4.1 achieved the highest zero-shot accuracy in both languages, scoring 83.09\% in Persian and 80.71\% in English. Notably, the best-performing open-weight model, DeepSeek-Chat-V3, achieved a strong 68.05\% in Persian and 73.30\% in English, followed closely by LLaMA-3.1-405B-Instruct with 67.02\% in Persian and 73.49\% in English. In terms of medical-tuned models, Meditron3-8B scored only 38.67\% in Persian and 50.00\% in English, revealing substantial room for improvement in domain adaptation for Persian.

Persian LLMs significantly underperformed across the board; some of them suffered greatly from poor instruction-following ability. PersianMind-1.0 achieved only 24.22\% in Persian (roughly equivalent to random guessing) and 25.17\% in English, suggesting limited medical knowledge and inadequate generalization capability in clinical domains. Similarly, Dorna2-LLaMA-3.1-8B-Instruct, another Persian LLM, scored just 34.87\% in Persian and 51.24\% in English, indicating slightly better instruction following but still poor domain alignment in the Persian medical setting.

Overall, closed-weight models consistently outperformed both open-weight and medical LLMs, particularly in Persian. While most models exhibited performance degradation when evaluated in Persian compared to English, some top-tier models — such as GPT-4.1 (83.09\% Persian vs.\ 80.71\% English) and Gemini-2.5-Flash-Preview (82.37\% Persian vs.\ 79.09\% English) — actually scored higher on the original Persian questions. This is consistent with our translation analysis: the English set is machine-translated from Persian, and even high-quality translation introduces subtle semantic drift and loss of clinical context that can slightly disadvantage models on the translated version.

We further analyzed model performance across different medical specialties. Figure~\ref{fig:field-heatmap} presents a heatmap of accuracy scores for each model across all medical fields in the PersianMedQA dataset.

Several factors shaped model performance across medical subfields. For example, pharmacology questions, which hinge on factual recall rather than complex clinical reasoning, yielded the highest accuracies for most models. Likewise, non-clinical items (theoretical or basic-science questions) tended to be answered more accurately than clinical case scenarios, reflecting their relatively straightforward nature.

In contrast, performance dropped sharply in subfields such as surgery and medical statistics, which require complex reasoning, quantitative interpretation, and a deeper understanding of language-specific clinical guidelines and protocols. These findings show that factual recall alone is insufficient: robust medical QA calls for deeper reasoning and cultural grounding across subfields.

\subsection{Translation Impact}

English dominates both the web-scale corpora that power modern LLMs and the medical literature on which they are trained. This bilingual evaluation is crucial for understanding a key trade-off in multilingual medical AI: while translating questions into English may align them better with a model's core knowledge base, it risks erasing subtle clinical guidelines and cultural contexts unique to local practice. To quantify this effect, we translated the PersianMedQA dataset into English and compared model performance on the original Persian versus the translated English questions.

To better understand model behavior across languages, we analyzed performance based on whether questions were answered correctly in only Persian, only English, or both. As expected, most models performed better on the English translations---even Persian LLMs---reflecting their predominant exposure to English medical data during training. However, a substantial subset of questions (ranging from 3-10\% across models) was answered correctly only in the original Persian. 

Manual analysis showed these cases often involved crucial local context that is lost in translation. These included healthcare system-specific protocols where Iranian clinical guidelines differ from Western standards, population-specific clinical considerations like regional disease prevalence, and semantic drift where the precise meaning of Persian medical terms is altered. Figure~\ref{fig:category3_vaccination_protocol_example} illustrates a representative case where Iran-specific antibiotic protocols lead to correct answers only when questions are presented in Persian, as translation obscures regional resistance patterns and drug availability that differ from Western guidelines.
\begin{figure*}[!t]
\centering
\begin{tcolorbox}[
enhanced,
colback=mybg,
colframe=myborder2,
arc=3mm,
boxrule=1pt,
title={\small\bfseries Regional Clinical Protocol Example},
fonttitle=\small\bfseries\centering,
left=4pt, right=4pt, top=4pt, bottom=4pt,
width=\textwidth
]
% Row 1: Questions with Answers
\begin{minipage}[t]{0.48\textwidth}
\textbf{\textcolor{myaccent}{\small Persian Question:}}
\begin{tcolorbox}[
colback=mysoft,
colframe=myborder,
boxrule=0.5pt,
arc=2mm,
left=3pt, right=3pt, top=3pt, bottom=3pt,
height=4.5cm
]
{\scriptsize\begin{persian}
نوجوان 13 ساله ای را بدلیل گاز گرفتگی سگ به اورژانس آورده اند. علاوه بر اقدامات مربوط به مراقبت زخم و پیشگیری هاری، جهت پیشگیری از کزاز تجویز کدامیک از موارد زیر برای این بیمار لازم است؟

\begin{enumerate}[label=\textbf{.\arabic*}, itemsep=1pt, rightmargin=0pt]
\item ایمونوگلوبولین کزاز
\item واکسن کزاز-دیفتری
\item ایمونوگلوبولین کزاز + واکسن کزاز-دیفتری
\item فقط مراقبت زخم کافی است
\end{enumerate}
\end{persian}}
\hrule
\vspace{3pt}
\scriptsize
\textbf{Persian Model:} \textcolor{successcolor}{Option 3} \textcolor{successcolor}{✓}\\[2pt]
\textbf{Correct Answer:} \textcolor{successcolor}{\textbf{Option 3}}
\end{tcolorbox}
\end{minipage}%
\hfill%
\begin{minipage}[t]{0.48\textwidth}
\textbf{\textcolor{myaccent}{\small English Translation:}}
\begin{tcolorbox}[
colback=mybg2,
colframe=myaccent2,
boxrule=0.5pt,
arc=2mm,
left=3pt, right=3pt, top=3pt, bottom=3pt,
height=4.5cm
]
\scriptsize
A 13-year-old adolescent has been brought to the emergency department due to a dog bite. In addition to wound care and rabies prevention measures, which of the following should be administered for tetanus prevention?
\begin{enumerate}[label=\arabic*., itemsep=1pt, leftmargin=12pt, topsep=2pt]
\item Tetanus Immune Globulin (TIG)
\item Td vaccine (Tetanus-Diphtheria)
\item TIG + Td vaccine
\item Wound care alone is sufficient
\end{enumerate}
\hrule
\vspace{3pt}
\textbf{English Model:} \textcolor{errorcolor}{Option 2 (Td only)} \textcolor{errorcolor}{✗}\\[2pt]
\textbf{Correct Answer:} \textcolor{successcolor}{\textbf{Option 3 (TIG + Td vaccine)}}
\end{tcolorbox}
\end{minipage}
\vspace{4pt}
% Row 2: Clinical Analysis (full width)
\textbf{\textcolor{myaccent}{\small Clinical Analysis:}}
\begin{tcolorbox}[
colback=mylight2,
colframe=myborder2,
boxrule=0.5pt,
arc=2mm,
left=4pt, right=4pt, top=4pt, bottom=4pt
]
\scriptsize
\textbf{\textcolor{myaccent}{Pattern:}} \textcolor{warningcolor}{\textbf{Regional Vaccination Schedule Differences}}\\[3pt]
\textbf{\textcolor{myaccent}{Key Insight:}} The Iranian national vaccination schedule's timing differs from Western protocols. In Iran, the last routine DTP dose is typically administered at age 6, meaning a 13-year-old is 7 years post-vaccination. For a contaminated wound (dog bite) in this context:
\begin{itemize}[itemsep=2pt, leftmargin=15pt]
\item \textbf{Iranian Protocol:} Both TIG and Td booster recommended for tetanus-prone wounds when $>$5 years since last dose
\item \textbf{Western Guidelines:} Often recommend Td booster alone if $<$10 years and primary series complete
\end{itemize}
\end{tcolorbox}
\end{tcolorbox}
\caption{Example demonstrating regional vaccination protocol differences affecting tetanus post-exposure prophylaxis decisions.}
\label{fig:category3_vaccination_protocol_example}
\end{figure*}

Such translation errors were most damaging in fields like surgery, where imprecise terminology led to incorrect answers even when the underlying reasoning was sound (see Appendix~\ref{appendix:cross_linguistic} for further examples).
\subsection{Impact of Model Size}

Our analysis of model size versus performance revealed that scale is not a universal solution. While larger general-purpose models like GPT-4.1 (83\% accuracy) clearly outperformed their smaller counterparts, increased size offered no advantage for specialized models. Both large medical-specific (e.g., MedAlpaca-13B) and Persian-tuned (e.g., Dorna2-LLaMA-3.1-8B) models struggled significantly, often scoring below smaller general-purpose models. These results underscore that model scale must be paired with high-quality, domain-relevant training data to achieve strong performance. 

\subsection{Prompting Strategies and Few-shot Learning}

We experimented with various prompting strategies and few-shot learning approaches; the results are summarized below.

\textbf{Role-based prompting}, where the model was instructed to act as a specialist based on the medical field of the question (e.g., "You are a cardiologist..."), resulted in slightly improved performance, but the gains were marginal.

\textbf{Few-shot learning}: For every test question, we drew the in-context examples exclusively from the PersianMedQA training split (up to k = 5 per query). We experimented with several retrieval schemes for picking those training examples: LaBSE cosine similarity, TF-IDF, and random selection, but none of them produced consistent gains over the zero-shot baseline. A plausible reason is the absence of high-quality embedding models tailored to Persian medical text, which makes it difficult to retrieve truly helpful training examples.

We also experimented with augmenting each question with a medical dictionary, extracted by a larger, more capable model (Gemini-2.5-Flash), that provided both translations and concise definitions of key terms. This dictionary is released alongside the dataset to help smaller models interpret domain-specific terminology. However, we found that this augmentation had a negligible effect on overall performance, especially for weaker or instruction-tuned models.

\subsection{Answer-Only Evaluation of LLM Medical Reasoning}

To test whether LLMs genuinely understand medical questions or merely exploit statistical patterns in answer choices, we adopted the \emph{partial-input} protocol of \cite{balepur2024artifactsabductionllmsanswer}. Each model received \emph{only} the four answer options without the question stem. Gemini-2.5-Flash-Preview achieved 35.60\% accuracy, substantially outperforming random guessing (25\%).

Manual inspection revealed that models exploited recurrent answer-choice artifacts, particularly evident in medical ethics—the highest-performing field at 46.8\% accuracy. In ethics questions, models can infer correct answers through: (i) \emph{hierarchical ethical principles}, where options containing phrases like "patient autonomy," "informed consent," or "professional disclosure" signal standard bioethical frameworks; (ii) \emph{logically exclusive options}, where choices violating fundamental medical ethics (e.g., "withhold information from patient") can be discarded; and (iii) \emph{linguistic cues}, where options with formal ethical terminology indicate textbook-correct responses. These patterns suggest that high performance in ethics may reflect recognition of moral vocabularies rather than genuine ethical reasoning, suggesting that medical MCQ benchmarks may overstate LLM capabilities by permitting exploitation of answer-choice artifacts.

\subsection{Model Ensembling}

Different models exhibit varied strengths across medical subjects, suggesting that ensembling diverse models can improve accuracy. Since top-performing models like GPT and Gemini are not open-weight, developing open-weight ensembles remains highly valuable.

\begin{table}[h!]
    \centering
    \caption{Majority-vote ensembles. ``$\Delta_{\text{best}}$'' is the gain over the best single model in the group.}
    \label{tab:ensemble-results}
    \scriptsize
    \begin{tabularx}{\linewidth}{lccc}
        \toprule
        \textbf{Ensemble / Baseline} & \textbf{Acc.} & \textbf{Avg. Acc.} & $\Delta_{\text{best}}$\\
        \midrule
        Top-3 Overall        & 0.834 & 0.808 & +0.003 \\
        Top-5 Overall        & 0.831 & 0.790 & -0.001 \\
        Top-3 GPT Family       & 0.803 & 0.704 & -0.028 \\
        Top-3 Google Family  & 0.795 & 0.728 & -0.029 \\
        Top-3 Claude Family  & 0.777 & 0.684 & -0.001 \\
        Top-5 open-weights   & \textbf{0.737} & 0.679 & \textbf{+0.033} \\
        \midrule
        Human Baseline       & 0.75 & — & — \\
        \bottomrule
    \end{tabularx}
\end{table}

As shown in Table \ref{tab:ensemble-results}, majority-vote ensembles of diverse model families outperformed single models, whereas same-family ensembles offer little benefit. Notably, an ensemble of five open-weight models achieved 73.7\% accuracy, a significant gain over the best individual model in that group (+0.033). 

\subsection{Chain-of-Thought Evaluation}

We evaluated the impact of Chain-of-Thought (CoT) prompting on four models: two large general-purpose models (GPT-4.1, Gemini 2.5 Flash), one medical model (Meditron3), and one Persian-language model (Dorna).

\textbf{Performance Gains}: For large general-purpose models, CoT improved accuracy by approximately 2\%, with the greatest gains on clinical questions, highlighting that clinical scenarios particularly benefit from explicit reasoning steps. Smaller models showed negligible improvement, likely due to weaker language and reasoning capabilities.

\textbf{Expert Analysis of CoT Errors}: A clinical expert reviewed GPT-4.1's CoT responses and identified four primary error types:
\begin{itemize}
    \item \textbf{Contextual Mismatch:} Applying reasoning based on non-Iranian clinical protocols.
    \vspace{-5 pt}
    \item \textbf{Ambiguity in Options:} Failing to distinguish between very similar or misleading answer choices.
    \vspace{-5 pt}
    \item \textbf{Reasoning Failures:} Exhibiting illogical or incomplete reasoning despite possessing the required knowledge.
    \vspace{-5 pt}
    \item \textbf{Knowledge Gaps:} Lacking the necessary factual information to answer correctly.
\end{itemize}

Examples for each category are available in Appendix~\ref{sec:appendix-errors}.

\section{Conclusion}
\label{sec:conclusion}

In this study, we present PersianMedQA, a dataset of 20,785 expert-validated Persian medical questions from 14 years of Iranian national medical exams, designed to evaluate how well current language models understand medical content across Persian and English contexts.

Our evaluation of 41 models revealed a significant performance hierarchy: closed-weight models like GPT-4.1 (83.1\% Persian, 80.7\% English) substantially outperform open-weight alternatives, with the best open-weight model (LLaMA-3.1-405B) achieving 67.0\% in Persian. Persian LLMs performed poorly (Dorna: 34.9\%), while medical LLMs showed only modest improvements over general models. Critically, our cross-linguistic analysis revealed that 3-10\% of questions require Persian-specific cultural and clinical knowledge, demonstrating that simple translation-based evaluation approaches are inadequate for medical AI in non-English contexts.

Future work should focus on developing Persian medical corpora for domain-specific training, creating retrieval-augmented systems that can access culturally appropriate medical guidelines, and expanding evaluation to other low-resource languages and multimodal medical contexts. Additionally, the training split of PersianMedQA remains underutilized in this study; future work should explore fine-tuning open-weight models directly on this split and compare against the prompt-based baselines established here. This work establishes a foundation for culturally grounded medical AI evaluation beyond English-centric benchmarks.

\section{Limitations}

Several factors constrained this study. (i) \emph{API restrictions and computational resources:} Cost and rate limits for commercial LLMs (e.g., GPT-4.1) reduced the number of evaluation runs and chain-of-thought variants we could conduct. Additionally, the lack of access to large-scale GPU infrastructure precluded fine-tuning experiments on the training split, which we leave for future work. (ii) \emph{Licensing barriers:} copyright restrictions prevented us from using larger multilingual biomedical corpora, limiting the scope of our experiments. As a result, our reported scores should be considered conservative lower bounds; broader data access and greater computational resources would enable a more exhaustive evaluation. (iii) \emph{Generalization scope:} The exam-based format of PersianMedQA is best suited for evaluating physician-level clinical reasoning. Results may not generalize to patient-focused question answering, such as informal health queries collected from online forums or patient communities, which represent a distinct and complementary evaluation setting.

\section{Ethics Statement}
This study involved the analysis and evaluation of LLMs on publicly available medical examination data. The dataset consists exclusively of multiple-choice examination questions from official Iranian medical licensing exams and contains no patient records, clinical notes, or any personally identifiable information. No anonymization was required, as the data is entirely de-identified by nature — all questions are impersonal exam items testing medical knowledge, not individual cases.

Regarding intellectual property, the questions are sourced from official Iranian national medical residency examinations administered by Sanjeshp (the Medical Education Assessment Center), a governmental body under the Iranian Ministry of Health. These examination papers are publicly released after each examination cycle and have been widely circulated by medical education institutions, test-preparation resources, and commercial study materials for over a decade. The dataset is used exclusively for non-commercial academic research purposes, consistent with established practices in the Persian NLP research community.

PersianMedQA is publicly available to support reproducible research and further development of Persian medical NLP. The dataset and bilingual medical dictionary are available at \href{https://huggingface.co/datasets/MohammadJRanjbar/PersianMedQA}{PersianMedQA}.

Our findings aim to improve the responsible deployment of language models in healthcare, especially for underrepresented languages. We emphasize that the models tested are not certified for clinical use and should not be deployed in real-world healthcare settings without strict oversight. We advocate for continued expert-in-the-loop development and further inclusion of diverse linguistic and cultural considerations in medical AI research.

\label{sec:reference}

\bibliographystyle{lrec2026-natbib}
\bibliography{lrec2026-example}

\newpage

\newpage

\section{Medical Specialist Background}
\label{appendix:specialist}

The medical specialist involved in this study is a board-certified internal medicine physician. She graduated from the University of Tehran with a degree in general medicine and completed her specialty training in internal medicine at Shahid Beheshti University of Medical Sciences. She has 5 years of clinical practice and has successfully passed the Iranian national medical licensing examinations. Her contributions to this work include validating subject classification, checking translation accuracy, overseeing the dataset curation process, and performing chain-of-thought (CoT) error analysis.

\section{Data Verification and Quality Assurance}
\label{appendix:data_verification}
\subsection{Answer Verification Process}
As described in Section~\ref{sec:construction}, all questions underwent a rigorous three-level verification process by the National Center for Medical Education Assessment (Sanjesh): (1) Initial expert committee review by board-certified medical professionals, (2) Public comment period where medical students and practitioners can report concerns or discrepancies, and (3) Final review and correction process incorporating community feedback. This established process has ensured high accuracy across 14 years of national medical examinations, though formal inter-annotator agreement scores in the traditional sense were not applicable given the institutional verification structure.
\subsection{Subject Classification Validation}
Subject classification was validated by the board-certified medical specialist described in Appendix~\ref{appendix:specialist}. The specialist reviewed ambiguous cases using a custom Telegram-based annotation interface (see Appendix~\ref{appendix:ui}), achieving greater than 90\% agreement with Gemini 2.5-Flash classifications over a randomly selected sample of 200 questions. We acknowledge that the absence of a second human annotator means formal inter-annotator agreement scores could not be computed; future work should include dual annotation on a representative subset to formally assess metadata quality.
\subsection{Question Diversity Distribution}
The National Center for Medical Education Assessment maintains natural diversity across medical specialties following standardized distributions mandated by the Ministry of Health. Each exam is dominated by the core clinical specialties: Pediatrics ($\sim$26 questions), Surgery ($\sim$24 questions), and Obstetrics/Gynecology ($\sim$18 questions), followed by Infectious Disease ($\sim$10 questions) and Pathology ($\sim$9 questions). The remaining 18 specialties—including Cardiology, Neurology, Psychiatry, Orthopedics, Endocrinology, Dermatology, and others—each contribute between 2 and 8 questions per exam, following official Ministry of Health guidelines.

\section{Examples of CoT Error Patterns}
\label{sec:appendix-errors}

This section presents representative error patterns identified in model-generated CoT outputs, as annotated by our clinical expert. For each example, we highlight the clinical context, the correct answer, the model's response, and a summary of the expert's evaluation.

\begin{tcolorbox}[enhanced, colback=teal!4!white, colframe=teal!80!black, arc=2mm, fonttitle=\bfseries, title={1. Contextual Mismatch}, boxrule=0.7pt, left=2mm, right=2mm, top=1mm, bottom=1mm]
	\scriptsize
	\textbf{Question:} What is the next step in an immunocompromised patient with nasal congestion and suspected invasive fungal sinusitis?\\[0.3em]
	\textbf{Correct Answer:} Endoscopy and biopsy\\[0.3em]
	\textbf{Model Answer:} Imaging (MRI) is needed before biopsy.\\[0.3em]
	\textbf{\textcolor{teal!80!black}{Expert Evaluation:}} The model follows a Western protocol; however, local clinical practice requires urgent biopsy due to high mortality risk.
\end{tcolorbox}

\vspace{0.3em}

\begin{tcolorbox}[enhanced, colback=cyan!3!white, colframe=cyan!60!black, arc=2mm, fonttitle=\bfseries, title={2. Ambiguity in Options}, boxrule=0.7pt, left=2mm, right=2mm, top=1mm, bottom=1mm]
	\scriptsize
	\textbf{Question:} What is the most common malignant neoplasm of the liver?\\[0.3em]
	\textbf{Correct Answer:} Hepatocellular carcinoma (HCC)\\[0.3em]
	\textbf{Model Answer:} Metastasis is more common overall, so we choose that.\\[0.3em]
	\textbf{\textcolor{cyan!70!black}{Expert Evaluation:}} The model selected a technically true but contextually incorrect answer; expert notes ambiguity in phrasing and clinical intent.
\end{tcolorbox}

\vspace{0.3em}

\begin{tcolorbox}[enhanced, colback=purple!3!white, colframe=purple!60!black, arc=2mm, fonttitle=\bfseries, title={3. Reasoning Failure}, boxrule=0.7pt, left=2mm, right=2mm, top=1mm, bottom=1mm]
	\scriptsize
	\textbf{Question:} What is the correct order of action in a 25-year-old with lymphoma and meningitis signs but no neurologic deficits?\\[0.3em]
	\textbf{Correct Answer:} Blood culture $\rightarrow$ Lumbar puncture $\rightarrow$ Empiric antibiotics\\[0.3em]
	\textbf{Model Answer:} CT scan should be done first due to immunosuppression.\\[0.3em]
	\textbf{\textcolor{purple!70!black}{Expert Evaluation:}} The patient's immunosuppression requires a different clinical approach, which the model failed to identify.
\end{tcolorbox}

\vspace{0.3em}

\begin{tcolorbox}[enhanced, colback=gray!3!white, colframe=gray!60!black, arc=2mm, fonttitle=\bfseries, title={4. Knowledge Gap}, boxrule=0.7pt, left=2mm, right=2mm, top=1mm, bottom=1mm]
	\scriptsize
	\textbf{Question:} Which drug works via motilin receptor stimulation for gastroparesis?\\[0.3em]
	\textbf{Correct Answer:} Erythromycin\\[0.3em]
	\textbf{Model Answer:} Metoclopramide is commonly used for gastroparesis, so we chose that.\\[0.3em]
	\textbf{\textcolor{gray!60!black}{Expert Evaluation:}} Model lacks pharmacologic mechanism knowledge and defaults to common treatments.
\end{tcolorbox}

\section{Few-shot Evaluation Prompt}
\label{appendix:fewshot_prompt}

In-context examples are drawn from the PersianMedQA training split using LaBSE cosine similarity, TF-IDF, and random selection (up to $k=5$).

\begin{tcolorbox}[
	breakable,
	enhanced,
	colback=mybg,
	colframe=myborder2,
	arc=3mm,
	boxrule=1pt,
	title={\scriptsize\bfseries Few-shot Prompt},
	fonttitle=\scriptsize\bfseries\centering,
	left=4pt, right=4pt, top=2pt, bottom=2pt
	]
	\scriptsize
	You are a medical expert tasked with answering multiple-choice medical questions.
	\vspace{0.2cm}
	
	\textbf{\textcolor{myaccent}{In-context Examples:}}
	\begin{tcolorbox}[
		colback=mylight2,
		colframe=myaccent2,
		boxrule=0.5pt,
		arc=2mm,
		left=3pt, right=3pt, top=2pt, bottom=2pt
		]
		\scriptsize
		\begin{verbatim}
			Question: [Example question 1]
			1: [Option 1]  2: [Option 2]
			3: [Option 3]  4: [Option 4]
			Answer: [Correct option number]
			...
			Question: [Example question k]
			1: [Option 1]  2: [Option 2]
			3: [Option 3]  4: [Option 4]
			Answer: [Correct option number]
		\end{verbatim}
	\end{tcolorbox}
	\vspace{0.2cm}
	
	\textbf{\textcolor{myaccent}{Now answer this question:}}
	\begin{tcolorbox}[
		colback=mylight2,
		colframe=myaccent2,
		boxrule=0.5pt,
		arc=2mm,
		left=3pt, right=3pt, top=2pt, bottom=2pt
		]
		\scriptsize
		\begin{verbatim}
			Question: [Medical question text]
			1: [Option 1]  2: [Option 2]
			3: [Option 3]  4: [Option 4]
		\end{verbatim}
	\end{tcolorbox}
	\vspace{0.2cm}
	
	\textbf{\textcolor{myaccent}{Important Notes:}}
	\begin{itemize}[nosep, leftmargin=10pt]
		\item Your output must be \textbf{only the option number} (1, 2, 3, or 4).
		\item Do \textbf{not} add explanations or extra text.
		\item Base your answers on authoritative medical knowledge.
	\end{itemize}
\end{tcolorbox}

\section{CoT Reasoning Prompt}
\label{appendix:cot_prompt}

\begin{tcolorbox}[
	breakable,
	enhanced,
	colback=mybg,
	colframe=myborder2,
	arc=3mm,
	boxrule=1pt,
	title={\scriptsize\bfseries CoT Prompt},
	fonttitle=\scriptsize\bfseries\centering,
	left=4pt, right=4pt, top=4pt, bottom=4pt
	]
	\scriptsize
	You are a medical expert taking a medical board examination.
	
	\vspace{0.3cm}
	\textbf{\textcolor{myaccent}{For each question, please:}}
	\begin{enumerate}[nosep, leftmargin=12pt]
		\item Read and understand the question carefully.
		\item Analyze the options (1--4) systematically.
		\item Apply your medical knowledge step by step.
		\item Show your chain-of-thought (CoT) reasoning clearly.
		\item Explain why each incorrect option is wrong, and the chosen one is correct.
		\item Explicitly state which option (1, 2, 3, or 4) is your final answer.
	\end{enumerate}
	
	\vspace{0.3cm}
	\textbf{\textcolor{myaccent}{Response format (JSON):}}
	\begin{tcolorbox}[
		colback=mylight2,
		colframe=myaccent2,
		boxrule=0.5pt,
		arc=2mm,
		left=3pt, right=3pt, top=3pt, bottom=3pt
		]
		\scriptsize
		\begin{verbatim}
			{
				"CoT": "your step-by-step reasoning",
				"Final_Answer": 1 | 2 | 3 | 4,
				"Reasoning": "concise justification"
			}
		\end{verbatim}
	\end{tcolorbox}
	
	\vspace{0.2cm}
	Be methodical, precise, and thorough in your analysis. Your expertise as \texttt{\{english\_specialty\}} is critical for answering these specialized questions correctly.
\end{tcolorbox}

\section{User Interfaces}
\label{appendix:ui}

To facilitate expert interaction throughout various phases of our study, we developed multiple Telegram-based interfaces to streamline collaboration with our medical specialist.

\subsection{Subject Annotation Interface}
We created a Telegram annotation bot to support subject-level classification. The specialist could review ambiguous or unclassified questions and select the most appropriate medical field from a predefined list of 23 specialties.

\begin{figure}[h]
	\centering
	\includegraphics[width=0.42\textwidth]{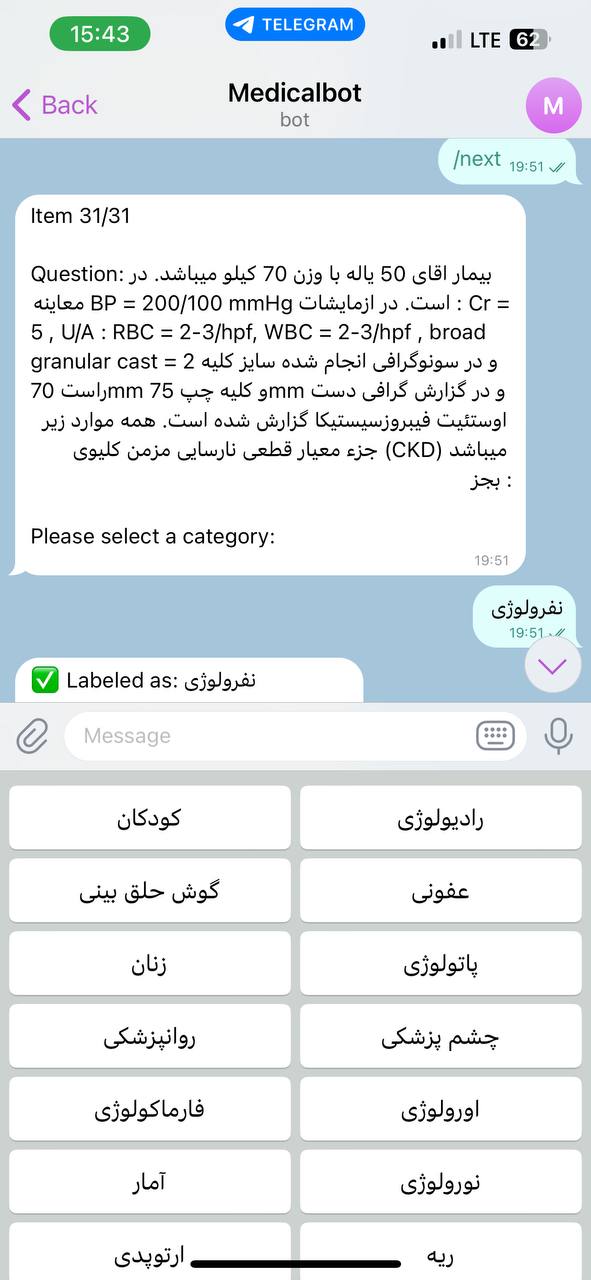}
	\caption{Telegram interface for expert subject classification of ambiguous questions.}
	\label{fig:subject_ui}
\end{figure}

\subsection{CoT Reasoning Interface}
To analyze the reasoning behind model outputs, we designed an interface that presented the expert with a curated 200-question subset. For each question, the expert was asked to: (1) select whether a predefined reasoning category applied (e.g., domain knowledge, commonsense, causal inference), (2) optionally assign a new category if none fit, and (3) provide a brief explanation justifying the correct answer.

\begin{figure}[h]
	\centering
	\includegraphics[width=0.42\textwidth]{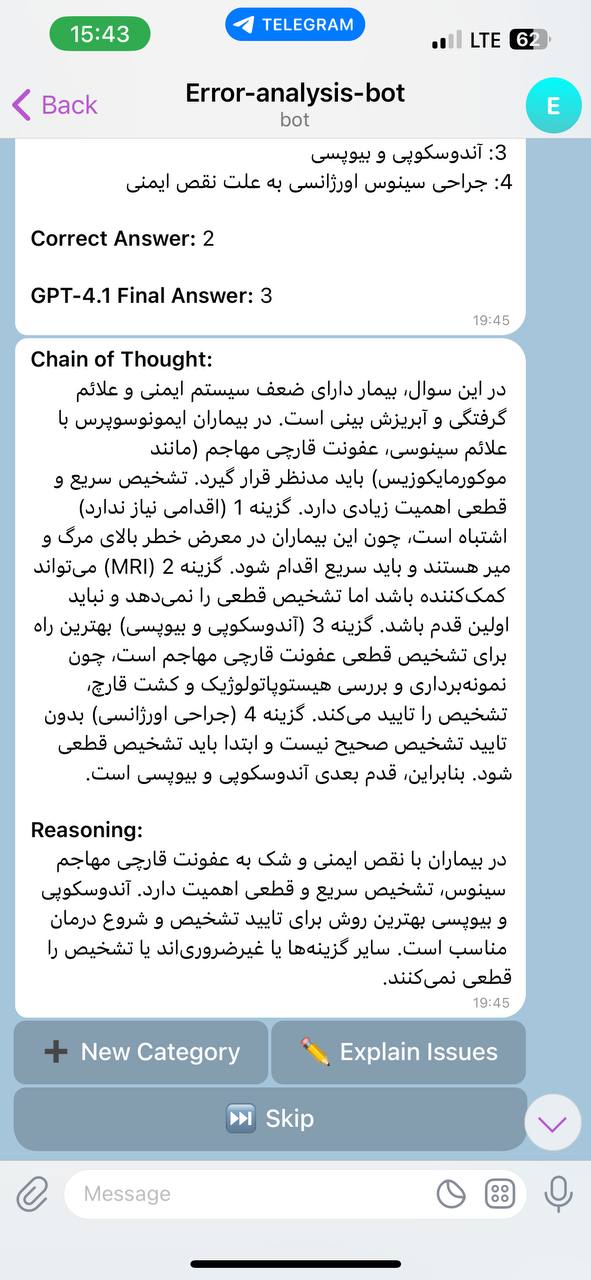}
	\caption{Telegram interface for expert annotation of reasoning categories and explanations.}
	\label{fig:reasoning_ui}
\end{figure}

\section{Persian Medical Dictionary}
\label{appendix:dictionary}

Table~\ref{tab:persian-medical-dictionary} summarizes the number of unique medical terms extracted per category in the bilingual Persian medical dictionary released alongside the dataset.

\begin{table}[h]
	\centering
	\caption{Distribution of extracted Persian medical terms.}
	\label{tab:persian-medical-dictionary}
	\scriptsize
	\begin{tabular}{lr}
		\toprule
		\textbf{Category} & \textbf{Unique Terms} \\
		\midrule
		Medical Devices          & 866    \\
		Medical Specialties      & 273    \\
		Lab Tests                & 6,410  \\
		Medical Abbreviations    & 2,596  \\
		Traditional Medicine Terms & 64   \\
		Procedures               & 9,632  \\
		Anatomical Terms         & 8,120  \\
		Symptoms                 & 14,397 \\
		Medications              & 5,905  \\
		Diseases                 & 16,400 \\
		\bottomrule
	\end{tabular}
\end{table}

\section{Overall Performance Comparison}
\label{appendix:overall-performance}

Table~\ref{tab:overall_performance} shows the zero-shot accuracy of all 41 evaluated models on the original Persian questions, the translated English questions, and their average. Models are sorted by average performance. The five lowest-performing models struggled significantly with instruction-following, resulting in accuracy scores at or below random guessing.

\newcommand{\tGC}{\textcolor{myaccent}{\textsuperscript{c}}}
\newcommand{\tGO}{\textcolor{myborder}{\textsuperscript{o}}}
\newcommand{\tM}{\textcolor{warningcolor}{\textsuperscript{m}}}
\newcommand{\tP}{\textcolor{errorcolor}{\textsuperscript{p}}}

\begin{table}[h]
	\centering
	\caption{Zero-shot accuracy of all evaluated models. \textcolor{myaccent}{\textsuperscript{c}}General Closed-Weight, \textcolor{myborder}{\textsuperscript{o}}General Open-Weight, \textcolor{warningcolor}{\textsuperscript{m}}Medical, \textcolor{errorcolor}{\textsuperscript{p}}Persian.}
	\label{tab:overall_performance}
	\scriptsize
	\begin{tabularx}{\linewidth}{Xccc}
		\toprule
		\textbf{Model} & \textbf{Fa (\%)} & \textbf{En (\%)} & \textbf{Avg (\%)} \\
		\midrule
		GPT-4.1\tGC                    & 83.09 & 80.71 & 81.90 \\
		Gemini-2.5-Flash-Preview\tGC   & 82.37 & 79.09 & 80.73 \\
		Claude-3.7-Sonnet\tGC          & 75.19 & 77.37 & 76.28 \\
		GPT-4.1-Mini\tGC               & 74.76 & 77.10 & 75.93 \\
		Gemini-2.0-Flash\tGC           & 76.86 & 74.50 & 75.68 \\
		DeepSeek-Chat-V3\tGO           & 68.05 & 73.30 & 70.67 \\
		LLaMA-3.1-405B-Instruct\tGO    & 67.02 & 73.49 & 70.25 \\
		LLaMA-4-Maverick\tGO           & 66.79 & 71.75 & 69.27 \\
		LLaMA-3.3-70B-Instruct\tGO     & 66.63 & 68.96 & 67.80 \\
		Qwen-2.5-72B-Instruct\tGO      & 65.17 & 70.26 & 67.72 \\
		LLaMA-4-Scout\tGO              & 63.29 & 69.12 & 66.21 \\
		Mistral-Saba\tGC               & 61.85 & 63.04 & 62.45 \\
		Gemma-3-27B-IT\tGO             & 59.06 & 62.74 & 60.90 \\
		Claude-3.5-Haiku\tGC           & 57.16 & 61.94 & 59.55 \\
		GPT-4.1-Nano\tGC               & 51.32 & 64.59 & 57.95 \\
		Gemma-3-12B-IT\tGO             & 52.22 & 57.85 & 55.03 \\
		Qwen-2.5-7B-Instruct\tGO       & 39.99 & 58.29 & 49.14 \\
		Mixtral-8x22B-Instruct\tGO     & 36.78 & 60.85 & 48.82 \\
		Aya-Expanse-8B\tGO             & 40.60 & 49.58 & 45.09 \\
		Meditron3-8B\tM                & 38.67 & 50.00 & 44.34 \\
		Mistral-Nemo\tGO               & 36.23 & 51.64 & 43.94 \\
		Meditron3-Qwen2.5-7B\tM        & 37.62 & 50.06 & 43.84 \\
		Dorna2-LLaMA-3.1-8B\tP         & 34.87 & 51.24 & 43.06 \\
		Cohere-Command-R7B\tGC         & 38.77 & 45.84 & 42.30 \\
		Gemma-3-4B-IT\tGO              & 35.87 & 42.25 & 39.06 \\
		Mistral-7B-Instruct\tGO        & 28.74 & 47.44 & 38.09 \\
		LLaMA-3.2-3B-Instruct\tGO      & 29.43 & 45.13 & 37.28 \\
		Meditron3-Qwen2.5-14B\tM       & 20.51 & 53.36 & 36.94 \\
		LLaMA-3.1-8B-Instruct\tGO      & 30.85 & 41.46 & 36.16 \\
		Internistai\tM                 & 21.85 & 48.34 & 35.09 \\
		Meditron3-Gemma2-2B\tM         & 27.44 & 34.97 & 31.21 \\
		Medicine-LLM\tM                & 24.85 & 33.21 & 29.03 \\
		BioMistral-7B\tM               & 25.76 & 31.38 & 28.57 \\
		LLaMA-3.2-1B-Instruct\tGO      & 26.44 & 25.48 & 25.96 \\
		Aya-23-8B\tGO                  & 27.77 & 23.47 & 25.62 \\
		PersianMind-1.0\tP             & 24.22 & 25.17 & 24.69 \\
		MedAlpaca-7B\tM                & 15.18 & 20.38 & 17.78 \\
		Meditron-7B\tM                 &  3.28 &  5.90 &  4.59 \\
		Meditron3-Gemma2-9B\tM         &  2.41 &  5.39 &  3.90 \\
		MedAlpaca-13B\tM               &  1.41 &  2.16 &  1.79 \\
		PersianLLaMA-13B\tP            &  0.00 &  0.00 &  0.00 \\
		\bottomrule
	\end{tabularx}
\end{table}

\section{Cross-Linguistic Performance Analysis}
\label{appendix:cross_linguistic}

Our cross-linguistic analysis revealed three distinct performance patterns across models. Representative examples for each category are provided below.

\subsection*{Category 1: Correct in Both Languages}
These questions involve standardized clinical protocols and universal pathophysiological concepts that transfer seamlessly across languages.

\begin{figure}[h!]
	\centering
	\begin{tcolorbox}[
		enhanced,
		colback=mybg,
		colframe=myborder2,
		arc=3mm,
		boxrule=1pt,
		title={\scriptsize\bfseries Category 1: Emergency Management Protocol},
		fonttitle=\scriptsize\bfseries\centering,
		left=3pt, right=3pt, top=3pt, bottom=3pt
		]
		\begin{minipage}[t]{0.47\linewidth}
			\textbf{\textcolor{myaccent}{\tiny English:}}
			\begin{tcolorbox}[colback=mybg2, colframe=myaccent2, boxrule=0.5pt, arc=2mm, left=2pt, right=2pt, top=2pt, bottom=2pt, height=1.7cm]
				\tiny
				An elderly male patient is hospitalized due to acute biliary pancreatitis. After 24 hours, the patient is still ill. Symptoms of biliary obstruction are evident. What is the best course of action?
			\end{tcolorbox}
		\end{minipage}
		\hfill
		\begin{minipage}[t]{0.47\linewidth}
			\textbf{\textcolor{myaccent}{\tiny Persian:}}
			\begin{tcolorbox}[colback=mysoft, colframe=myborder, boxrule=0.5pt, arc=2mm, left=2pt, right=2pt, top=2pt, bottom=2pt, height=1.7cm]
				\tiny
				\begin{persianpar}
بیمار آقای مسنی به‌دلیل پانکراتیت صفراوی حاد بستری است. بعد از 24 ساعت درمان همچنان حال بیمار وخیم است. علائم انسداد صفراوی مشهود است. بهترین اقدام کدام است؟
				\end{persianpar}
			\end{tcolorbox}
		\end{minipage}
		\vspace{3pt}
		\begin{tcolorbox}[colback=mylight2, colframe=myborder2, boxrule=0.5pt, arc=2mm, left=2pt, right=2pt, top=2pt, bottom=2pt]
			\tiny
			\textbf{\textcolor{myaccent}{Pattern:}} \textcolor{successcolor}{\textbf{Universal Protocol}} — Emergency biliary obstruction management is consistent across healthcare systems, transferring correctly in both languages.
		\end{tcolorbox}
	\end{tcolorbox}
	\caption{\scriptsize Category 1: correct in both languages.}
	\label{fig:category1_example}
\end{figure}

\subsection*{Category 2: Correct Only After Translation}
These questions benefit from the model's stronger English medical training, particularly in specialized terminology.
\begin{figure}[h]
	\centering
	\begin{tcolorbox}[
		enhanced,
		colback=mybg,
		colframe=myborder2,
		arc=3mm,
		boxrule=1pt,
		title={\scriptsize\bfseries Category 2: Specialized Anatomical Pathology},
		fonttitle=\scriptsize\bfseries\centering,
		left=3pt, right=3pt, top=3pt, bottom=3pt
		]
		\begin{minipage}[t]{0.47\linewidth}
			\textbf{\textcolor{myaccent}{\tiny English:}}
			\begin{tcolorbox}[colback=mybg2, colframe=myaccent2, boxrule=0.5pt, arc=2mm, left=2pt, right=2pt, top=2pt, bottom=2pt, height=1.5cm]
				\tiny
				Which finger's flexor tendon sheath infection (infectious tenosynovitis) is at risk of spreading to the forearm?
			\end{tcolorbox}
		\end{minipage}
		\hfill
		\begin{minipage}[t]{0.47\linewidth}
			\textbf{\textcolor{myaccent}{\tiny Persian:}}
			\begin{tcolorbox}[colback=mysoft, colframe=myborder, boxrule=0.5pt, arc=2mm, left=2pt, right=2pt, top=2pt, bottom=2pt, height=1.5cm]
				\tiny
				\begin{persianpar}
					خطر گسترش تنوسینوویت عفونی تاندون فلکسور کدام انگشت دست به ساعد وجود دارد؟
				\end{persianpar}
			\end{tcolorbox}
		\end{minipage}
		\vspace{3pt}
		\begin{tcolorbox}[colback=mylight2, colframe=myborder2, boxrule=0.5pt, arc=2mm, left=2pt, right=2pt, top=2pt, bottom=2pt]
			\tiny
			\textbf{\textcolor{myaccent}{Pattern:}} \textcolor{infocolor}{\textbf{English Literature Dominance}} — Specialized anatomical terminology is predominantly represented in English medical literature, improving performance post-translation.
		\end{tcolorbox}
	\end{tcolorbox}
	\caption{\scriptsize Category 2: correct only after translation.}
	\label{fig:category2_example}
\end{figure}

\subsection*{Category 3: Correct Only in Persian}
These questions involve Iran-specific medical practices or clinical contexts that are altered or lost during translation.

\begin{figure}[h]
	\centering
	\begin{tcolorbox}[
		enhanced,
		colback=mybg,
		colframe=myborder2,
		arc=3mm,
		boxrule=1pt,
		title={\scriptsize\bfseries Category 3: Iran-Specific Clinical Protocols},
		fonttitle=\scriptsize\bfseries\centering,
		left=3pt, right=3pt, top=3pt, bottom=3pt
		]
		\textbf{\textcolor{myaccent}{\tiny Example A: Regional Antibiotic Protocols}}
		\vspace{2pt}
		
		\begin{minipage}[t]{0.47\linewidth}
			\begin{tcolorbox}[colback=mybg2, colframe=myaccent2, boxrule=0.5pt, arc=2mm, left=2pt, right=2pt, top=2pt, bottom=2pt, height=1.5cm]
				\tiny
				A 30-year-old motorcyclist presents with a $\sim$10 cm wound over the right tibia with a comminuted fracture. Which antibiotic and for how long?
			\end{tcolorbox}
		\end{minipage}
		\hfill
		\begin{minipage}[t]{0.47\linewidth}
			\begin{tcolorbox}[colback=mysoft, colframe=myborder, boxrule=0.5pt, arc=2mm, left=2pt, right=2pt, top=2pt, bottom=2pt, height=1.5cm]
				\tiny
				\begin{persianpar}
					موتورسوار 30 ساله ای با زخم 10 سانتی متری روی ساق پای راست به همراه شکستگی خرد شده تیبیا به اورژانس مراجعه می کند. از کدام آنتی بیوتیک و برای چه مدت؟
				\end{persianpar}
			\end{tcolorbox}
		\end{minipage}
		
		\vspace{4pt}
		\textbf{\textcolor{myaccent}{\tiny Example B: Vaccination Schedule Differences}}
		\vspace{2pt}
		
		\begin{minipage}[t]{0.47\linewidth}
			\begin{tcolorbox}[colback=mybg2, colframe=myaccent2, boxrule=0.5pt, arc=2mm, left=2pt, right=2pt, top=2pt, bottom=2pt, height=1.5cm]
				\tiny
				A 17-year-old presents with a contaminated wound. History indicates routine national vaccination has been performed.
			\end{tcolorbox}
		\end{minipage}
		\hfill
		\begin{minipage}[t]{0.47\linewidth}
			\begin{tcolorbox}[colback=mysoft, colframe=myborder, boxrule=0.5pt, arc=2mm, left=2pt, right=2pt, top=2pt, bottom=2pt, height=1.5cm]
				\tiny
				\begin{persianpar}
					نوجوان 17 ساله با زخم آلوده به خاک به اورژانس آمده. در سابقه واکسیناسیون روتین کشوری انجام داده است.
				\end{persianpar}
			\end{tcolorbox}
		\end{minipage}
		
		\vspace{3pt}
		\begin{tcolorbox}[colback=mylight2, colframe=myborder2, boxrule=0.5pt, arc=2mm, left=2pt, right=2pt, top=2pt, bottom=2pt]
			\tiny
			\textbf{\textcolor{myaccent}{Pattern:}} \textcolor{warningcolor}{\textbf{Regional Protocol Differences}} — Iranian antibiotic and vaccination protocols differ from Western guidelines. Critical clinical context is lost in translation, leading to incorrect answers on the English version.
		\end{tcolorbox}
	\end{tcolorbox}
	\caption{\scriptsize Category 3: correct only in Persian due to Iran-specific protocols.}
	\label{fig:category3_example}
\end{figure}

\end{document}